\documentclass[letterpaper]{article} 
\usepackage{aaai2027}  
\usepackage[hyphens]{url}  
\usepackage{graphicx} 
\usepackage{natbib}  
\usepackage{caption} 
\usepackage{algorithm}
\usepackage{algorithmic}

\usepackage{newfloat}
\usepackage{amsmath,amssymb}
\usepackage{xcolor}
\usepackage{placeins}
\usepackage{listings}
\DeclareCaptionStyle{ruled}{labelfont=normalfont,labelsep=colon,strut=off} 
\floatstyle{ruled}
\newfloat{listing}{tb}{lst}{}
\floatname{listing}{Listing}

\usepackage{booktabs}

\usepackage{xspace}
\newcommand{\method}{GraphIR\xspace}

\title{GraphIR: Architecture-Level Search States for LLM-Guided Neural Architecture Evolution}
\author{
    Zhen Liu\equalcontrib\textsuperscript{\rm 1},
    Wanqi Zhou\equalcontrib\textsuperscript{\rm 1},
    Shuanghao Bai\textsuperscript{\rm 1},\\
    Yuhan Liu\textsuperscript{\rm 2},
    Jinjun Wang\textsuperscript{\rm 1},
    Jingwen Fu\textsuperscript{\rm 3,\rm 4}\corresponding
}
\affiliations{
    \textsuperscript{\rm 1}National Key Laboratory of Human-Machine Hybrid Augmented Intelligence, and Institute of Artificial Intelligence and Robotics, Xi'an Jiaotong University\\
     \textsuperscript{\rm 2}MiLM Plus, Xiaomi Inc.\\
    \textsuperscript{\rm 3}Zhongguancun Academy\\
    \textsuperscript{\rm 4}Zhongguancun Institute of Artificial Intelligence\\
    fujingwen@bza.edu.cn

}

\begin{document}

\maketitle

\begin{abstract}
Large language models (LLMs) enable neural architecture search (NAS)
directly over executable neural network programs. However, code-level
flexibility does not provide the architecture state needed for effective
mutation: LLMs must infer tensor dependencies, editable components, and
compatibility constraints from implementation details. To address this
representation mismatch, we propose \method, an architecture-aware
intermediate representation that supplements executable programs with a
mutation-aligned candidate state. \method organizes each candidate
through three complementary views: a \emph{computation skeleton}
describing tensor flow, a \emph{mutation surface} exposing editable
modules and operations, and a \emph{validity envelope} capturing
interface contracts, propagated shapes, and downstream dependencies.
To evaluate our method, we construct
\emph{NAS-Dependency}, a 120-question benchmark covering six
complementary dependency-reasoning dimensions.
The diagnostic shows that \method is particularly effective at
identifying exact producer occurrences, tracing dependency propagation,
and diagnosing interface and failure risks.
Across six downstream benchmarks including CLRS, \method achieves the
best overall search performance while maintaining comparable model size
and favorable end-to-end NAS efficiency when integrated into
OpenEvolve.
These results show that a mutation-oriented architecture
state provides an effective interface between executable neural programs
and LLM-guided architecture evolution.

\end{abstract}

\section{Introduction}

Neural architecture design is fundamentally a problem of representation.
A candidate architecture is not merely an executable program, but a
structured computation whose modules, connections, tensor
transformations, and output heads jointly determine its inductive bias,
computational cost, and task performance. Neural architecture search
(NAS) automates this design process by exploring candidate architectures
under a task-specific evaluation budget
\cite{NAS2019survey,white2023neural}. Classical NAS methods make the
candidate representation explicit through 
\begin{center}
\begin{minipage}{\columnwidth}
    \centering
    \includegraphics[width=0.86\linewidth]{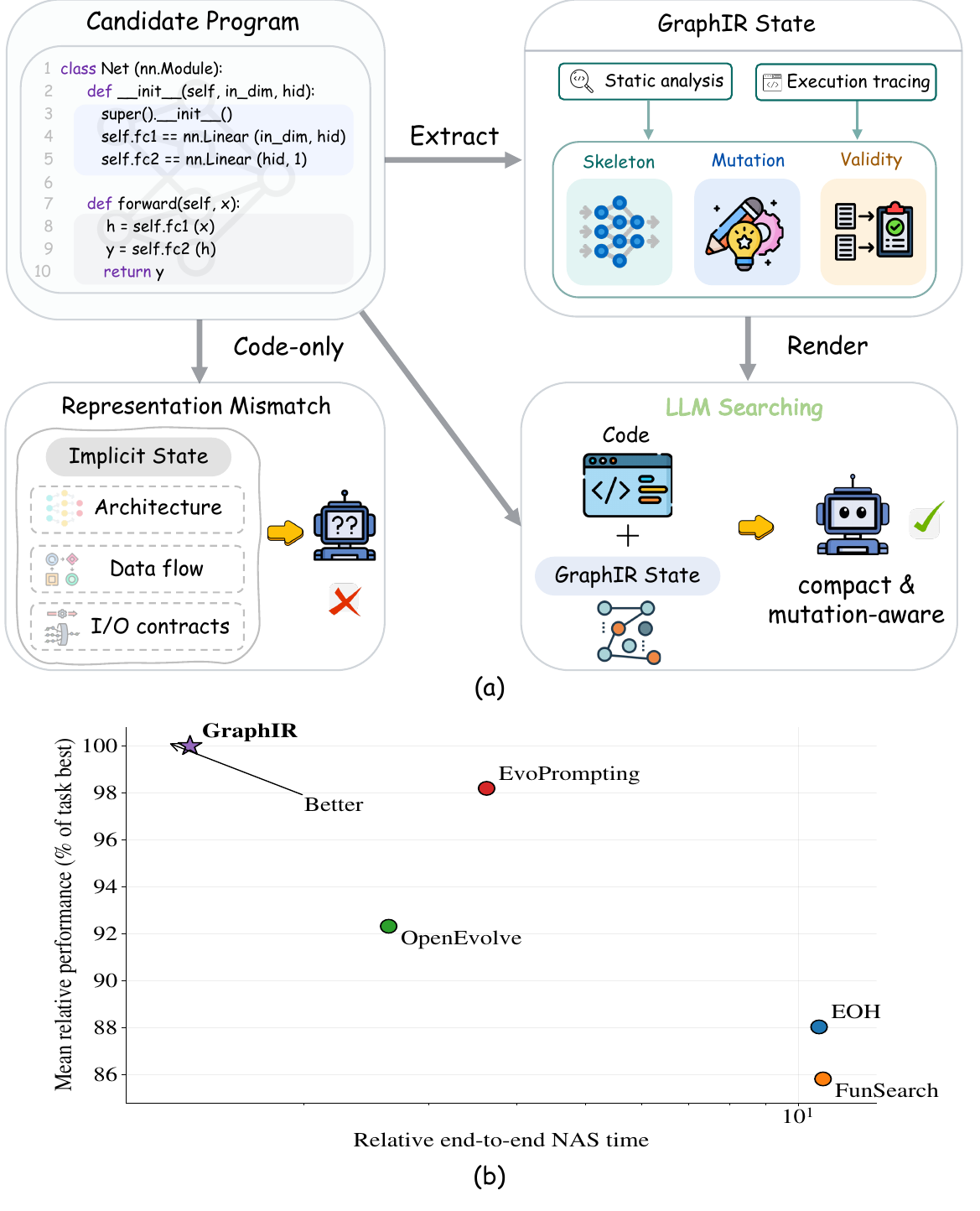}
    \captionof{figure}{Overview and evaluation of \method.
    \textbf{(a)} \method extracts a compact, mutation-aligned architecture
    state from executable neural network code, exposing its computation
    skeleton, mutation surface, and validity envelope for LLM-guided search.
    \textbf{(b)} Cross-task efficiency--performance comparison on 20News,
    CartPole-v1, Yeast, and Breast Cancer Wisconsin Diagnostic. Performance
    is normalized by the best result on each benchmark, while end-to-end NAS
    time is normalized by the fastest method and aggregated using the
    geometric mean. Points closer to the upper-left corner are preferred.
    Detailed experiment results are provided in the supplementary material.}
    \label{fig:graphir_overview}
\end{minipage}
\end{center}

cell graphs, operation sets,
hierarchical encodings, or architecture grammars, and optimize over this
representation using reinforcement learning
\cite{NASRL2017iclr}, evolutionary search
\cite{NASEN2019AAAI}, or differentiable optimization
\cite{differ2019iclr}. Such representations provide typed edit choices,
validity constraints, and clear mutation semantics, but restrict search
to a manually defined architecture space.

Recent LLM-guided NAS methods pursue a more open-ended alternative by
allowing large language models to propose architecture edits in code,
natural language, or graph form
\cite{evoprompt2023neurips,designaaai2025,liu2026structured}. In
code-based NAS, the candidate is an executable neural network program,
allowing the LLM to modify modules, branches, tensor operations,
normalization layers, residual paths, gating mechanisms, pooling
operations, and prediction heads through ordinary source-level edits.
This flexibility substantially expands the search space beyond a fixed
NAS grammar. However, it also introduces a representation mismatch:
executable code provides an open-ended mutation substrate, but leaves
the architecture state required for mutation largely implicit, as
illustrated in Figure~\ref{fig:graphir_overview}(a). 
This obscures mutation targets and dependencies.

The central difficulty is that source code is optimized for execution,
rather than architecture mutation. It interleaves architecture-level
decisions with implementation details such as variable names, helper
functions, tensor reshaping, module construction, and evaluator
interfaces. Consequently, the LLM must infer which components are
meaningful mutation targets, how tensor dimensions are coupled, which
downstream modules depend on an edited representation, and which
input-output contracts must remain unchanged. This burden can produce
syntactically valid but architecturally inconsistent candidates, for
example through incompatible tensor shapes or modified evaluator-facing
outputs. It can also produce executable but ineffective mutations that
change implementation details without introducing a meaningful
architectural variation.

This paper asks: \emph{What candidate state should an LLM observe when
mutating an executable neural architecture?} We argue that a
mutation-aligned state needs three complementary views. The
\emph{computation skeleton} traces tensor flow from inputs to outputs.
The \emph{mutation surface} identifies editable modules,
hyperparameters, and tensor operations. The \emph{validity envelope}
records input-output contracts, tensor shapes, dimension couplings, and
downstream dependencies. Together, they define what may change, what
must remain valid, and how local edits affect the architecture.

We therefore propose \method, an architecture-aware intermediate
representation for LLM-guided code-based NAS. \method preserves the
source program as the executable and editable search object, while
supplementing it with an explicit architecture-level candidate state.
Given a neural network program, \method combines static analysis and
execution tracing to extract architecture-relevant modules, tensor
operations, structural facts, input-output contracts, runtime dataflow,
and propagated shapes. The resulting state is rendered as compact
textual context and appended to the mutation prompt. As shown in
Figure~\ref{fig:graphir_overview}(a), the source code remains the primary
editing anchor, while \method provides complementary structural guidance
for architecture mutation.
This separation preserves open-ended code editing while making
architecture reasoning explicit.

\method differs from both generic program graphs and deep-learning
compiler intermediate representations. Abstract syntax trees,
control-flow graphs, data-flow graphs, program-dependence graphs, and
code property graphs are designed primarily for program-analysis
questions, such as variable definitions, control dependencies, and
function calls. Deep-learning IRs are typically designed for model
tracing, transformation, interchange, lowering, or execution
optimization. In contrast, \method targets mutation-specific questions:
which components constitute the architecture, which operations are
editable, which interfaces must remain valid, and which downstream
components are affected by a local edit. It is therefore not a
general-purpose code graph, but a mutation-aligned architecture state
for LLM-guided search.

To isolate this representation mechanism, we introduce
\emph{NAS-Dependency}, a 120-question benchmark covering six dimensions
of architecture-level dependency reasoning. Compared with code-only
prompts and alternative structured representations, \method achieves
the highest macro average and the best or tied-best result on four
dimensions, with its main advantages appearing in producer localization,
dependency propagation, and interface-failure diagnosis.

We further evaluate \method on six downstream architecture-search
benchmarks spanning five domains: CLRS for algorithmic reasoning,
MNIST1D-Shuffle for 1D sequence classification
\cite{greydanus2024scaling}, 20 Newsgroups for text classification
\cite{twenty_newsgroups_113}, CartPole-v1 for reinforcement-learning
control \cite{rlbench2024}, and Yeast \cite{yeast_110} and Breast Cancer
Wisconsin Diagnostic \cite{breast_cancer_wisconsin} for scientific
tabular classification. On the 30-task CLRS suite, \method achieves the
highest average accuracy of \(89.21\%\) while maintaining comparable
model size. Controlled MNIST1D-Shuffle experiments further show that
\method outperforms alternative structured representations in both
in-distribution and out-of-distribution accuracy. On the remaining four
benchmarks, it achieves the best task performance in every case and the
lowest aggregated relative end-to-end NAS time, as summarized in
Figure~\ref{fig:graphir_overview}(b). Together, these results demonstrate
that mutation-aligned architecture states improve both the reasoning
mechanism and downstream effectiveness of LLM-guided architecture
evolution.

Our contributions are as follows:
\begin{itemize}
\item We identify a representation mismatch in LLM-guided code-based
NAS: executable code supports open-ended mutation but does not explicitly
expose the architecture state required for effective editing.

\item We propose \method, a compact architecture-aware representation
that supplements executable neural programs with a computation skeleton,
mutation surface, and validity envelope.

\item We introduce \emph{NAS-Dependency} for architecture-level
reasoning and evaluate \method across six downstream benchmarks. The
results show consistent improvements in dependency reasoning, search
performance, and end-to-end NAS efficiency.
\end{itemize}

\section{Related Work}

\paragraph{Neural architecture search and candidate representations.}
NAS is commonly organized around three components: the search space, the search strategy, and the performance estimation strategy~\cite{NAS2019survey,white2023neural}. 
Classical NAS methods instantiate these components through reinforcement learning, evolutionary search, differentiable relaxation, Bayesian optimization, and predictor-guided search~\cite{NASRL2017iclr,NASEN2019AAAI,differ2019iclr,kandasamy2018nasbot,white2021bananas}. 
A common feature of these methods \cite{fu2019recognition,fu2023momentum,fu2024understanding,fu2026overcoming} is that the candidate representation is explicitly specified in advance, for example as operation choices on a cell graph, discrete architectural hyperparameters, depth-width configurations, or continuous architecture parameters. 
Such representations make search states, validity rules, and mutation operators well defined. 
However, this structure is obtained by constraining candidates to a manually designed search space. 
Our work studies the complementary setting where the candidate is an executable neural network program, and asks what architecture-level state should be exposed to an LLM when the search space is no longer closed by a predefined grammar.

\paragraph{Program-based and LLM-guided architecture evolution.}
Recent work \cite{liu2026essence, liu2026instruction} has explored the use of large language models for neural architecture design and evolution. 
EvoPrompting uses language models as adaptive mutation and crossover operators for code-level NAS~\cite{evoprompt2023neurips}, while LLMatic combines LLM-generated code variations with quality-diversity optimization~\cite{gecco2024acm}. 
Other works use LLMs to infer transferable design principles~\cite{designaaai2025}, coordinate multiple architecture-design agents with graph-based architecture representations~\cite{yang2025nader,liu2026elva}, encode architectures as unified numerical strings for cross-domain search~\cite{hu-etal-2025-lm}, or condition code edits on structured functional factors~\cite{liu2026structured}. 
These studies demonstrate that LLMs can inject useful architectural priors into NAS, but they also reveal a central representation issue: the effectiveness of LLM-guided search depends strongly on what candidate state is presented to the model. 
In contrast to methods that rely primarily on raw source code, natural-language principles, graph-only architectures, or numerical encodings, \method focuses on code-based NAS and supplements the editable program with an architecture-aware intermediate representation for mutation reasoning.

\paragraph{Graph-based architecture representations and structured code contexts.}
Graph-based representations have been used in NAS to encode candidate architectures for performance prediction or search acceleration. 
Representative methods such as NGE, GATES, and NASGEM represent neural architectures as graphs and learn architecture embeddings or graph similarities for predictor- or estimator-based NAS~\cite{li2020neural,ning2020generic,cheng2021nasgem}. 
In contrast, our goal is not to encode a predefined architecture graph into an embedding, but to extract mutation-relevant structure from executable neural network code.
Structured code representations have also been used to support LLM-based code understanding and generation, including data-flow-aware code models, code context graphs, graphical code retrieval, and repository-level code graphs~\cite{guo2021graphcodebert,liu2024graphcoder,du2024codegrag,ouyang2025repograph}. 
However, these representations target general program reasoning rather than neural architecture mutation. 
\method differs by constructing a tensor-aware architecture state that exposes modules, tensor operations, executable tensor dataflow, shapes, mutation sites, and compatibility constraints, enabling safe LLM-guided architecture modification beyond fixed NAS encodings.

\section{Problem Formulation}

We consider LLM-guided neural architecture evolution over executable neural network programs. At iteration $t$, the algorithm maintains an archive
\begin{equation}
    \mathcal{A}_t = \{(P_i, s_i)\}_{i=1}^{N_t},
\end{equation}
where $P_i$ denotes a previously evaluated program and $s_i$ is its score. A parent program $P_t^{\mathrm{par}}$ is selected from the programs in $\mathcal{A}_t$, and an LLM mutation operator proposes a child program:
\begin{equation}
    P_{t+1} \sim \pi_{\mathrm{LLM}}
    \left(\cdot \mid \mathcal{P}(P_t^{\mathrm{par}}, H_t)\right),
\end{equation}
where $H_t$ denotes evaluation feedback and $\mathcal{P}(\cdot)$ is the mutation prompt. The child program is then validated, evaluated, and inserted into the archive according to the update rule.

Unlike conventional NAS, where mutation usually operates on a predefined architecture encoding and then instantiates executable code, LLM-guided code evolution directly mutates the source program. Although $P_t^{\mathrm{par}}$ implicitly contains the architecture, its NAS-relevant structure is not explicitly exposed to the LLM. We therefore introduce an extractor
\begin{equation}
    z_t = \phi(P_t^{\mathrm{par}}),
\end{equation}
where $z_t$ is the architecture-level state rendered by \method. It summarizes information such as modules, execution flow, mutable components, and interface constraints. The mutation process is then augmented as
\begin{equation}
    P_{t+1} \sim \pi_{\mathrm{LLM}}
    \left(\cdot \mid \mathcal{P}(P_t^{\mathrm{par}}, z_t, H_t)\right).
\end{equation}
Here, $z_t$ supplements the source code with explicit architecture-level information for valid and useful mutation.

\section{Method}

\subsection{Overview}

LLM-guided code-based NAS searches directly over executable neural
network programs. This program space permits open-ended architectural
modifications, but the architecture state of a candidate remains
implicit in implementation-level code. Before proposing a mutation, the
LLM must therefore infer the model interface, architecture components,
tensor transformations, and dataflow dependencies from source syntax.

We introduce \method to expose this information as an explicit
architecture-level candidate state. Given a parent program
$P_t^{\mathrm{par}}$, \method constructs
\begin{equation}
    G_t =
    \operatorname{GraphIR}
    \left(
        P_t^{\mathrm{par}}
    \right),
    \label{eq:graphir-extraction}
\end{equation}
where $G_t$ contains candidate-specific architecture evidence obtained
from complementary source-level and execution-grounded analyses. The
source program remains the executable implementation to be modified,
while $G_t$ provides architecture-level context for mutation. The child
program is generated according to
\begin{equation}
    P_{t+1}
    \sim
    \pi_{\mathrm{LLM}}
    \left(
        \cdot
        \mid
        \mathcal{P}
        \left(
            P_t^{\mathrm{par}},
            G_t,
            H_t
        \right)
    \right),
    \label{eq:graphir-guided-mutation}
\end{equation}
where $H_t$ denotes the available evaluation history and
$\mathcal{P}$ denotes the mutation-prompt construction procedure.
GraphIR is reconstructed for every generated candidate before that
candidate participates in subsequent search iterations.

\subsection{GraphIR State Construction}

Given a candidate program $P$, \method constructs an LLM-visible
architecture state
\begin{equation}
    G(P)
    =
    \left(
        \mathcal{I},
        \mathcal{O},
        \mathcal{M},
        \mathcal{T},
        \mathcal{F},
        \mathcal{E}
    \right),
    \label{eq:graphir-state}
\end{equation}
where $\mathcal{I}$ and $\mathcal{O}$ denote the input and output
interface contracts, $\mathcal{M}$ contains architecture modules,
$\mathcal{T}$ contains architecture-relevant tensor operations,
$\mathcal{F}$ contains deterministic structure facts, and
$\mathcal{E}$ contains selected producer-consumer dependencies.

These fields provide three complementary views of the candidate. The
\emph{computation skeleton}, represented by
$(\mathcal{M},\mathcal{T},\mathcal{E})$, describes the main tensor
transformations and their ordering. The \emph{mutation surface},
represented by $(\mathcal{M},\mathcal{T},\mathcal{F})$, exposes
architecture components and structural properties that may serve as
grounded mutation targets. The \emph{validity envelope}, represented by
$(\mathcal{I},\mathcal{O},\mathcal{E})$, records the external interface
and internal dependencies that should remain compatible after mutation.

The contracts $\mathcal{I}$ and $\mathcal{O}$ are constructed in two
stages. GraphIR first classifies the candidate using deterministic
source patterns and initializes a profile-specific default contract.
Programs containing a registered graph-processor factory, or the
characteristic combination of node features, edge features, and an
adjacency matrix, receive a graph-structured contract; other candidates
receive the generic PyTorch contract. GraphIR then instantiates the
candidate and synthesizes lightweight example inputs. For a standard
model-construction interface, the input width of the first linear module
is used to construct an example tensor; a predefined feature width is
used when this dimension cannot be inferred. PyTorch FX symbolic
tracing~\cite{reed2022torchfx} recovers the computation graph, and shape
propagation annotates its nodes. Whenever observed input or output
shapes are available, they independently replace the corresponding
default contracts. 

The module set $\mathcal{M}$ and operation set $\mathcal{T}$ are
obtained from the editable program region using Python abstract syntax
tree analysis. Constructor calls in assignment statements are matched
against a fixed neural-module vocabulary, including linear,
convolutional, normalization, recurrent, attention, regularization, and
pooling modules. Function calls, tensor methods, and selected
expressions are matched against a fixed operation vocabulary covering
activations, shape transformations, concatenation, stacking,
reductions, masking, matrix multiplication, and tensor contraction.
Binary addition, loops, and conditional statements are additionally
recorded as architecture-relevant operations.

The structure facts $\mathcal{F}$ are generated through fixed
trigger-to-label rules rather than LLM summarization. Source-level rules
map recognized modules, operations, and lexical patterns to structural
annotations. For example, convolutional modules yield a
convolution/local-mixing fact, recurrent modules yield a state-update
fact, and explicit addition yields an additive-merge fact.
Execution-grounded rules similarly inspect the traced FX graph to
identify observed linear, convolutional, recurrent, shape-transform,
and additive-merge operations. The resulting source-level and
execution-grounded facts are merged and deduplicated.

The dependency set $\mathcal{E}$ prioritizes producer-consumer edges
recovered from the traced FX graph. If executable tracing is
unavailable, GraphIR uses conservative assignment-level dependencies
obtained from static analysis, where symbols referenced on the
right-hand side of an assignment are connected to its target.
Module-constructor dependencies and non-tensor attribute initialization
are excluded to reduce spurious edges.

Algorithm~\ref{alg:graphir} summarizes the construction procedure.

\begin{algorithm}[t]
\caption{GraphIR State Construction}
\label{alg:graphir}
\begin{algorithmic}[1]
\REQUIRE Candidate program $P$
\ENSURE GraphIR state $G(P)$ and rendered context $C(P)$
\STATE Select the marked evolution region, or use the complete program
\STATE Initialize profile-specific contracts $\mathcal{I}$ and $\mathcal{O}$
\STATE Extract $\mathcal{M}$, $\mathcal{T}$, static facts
$\mathcal{F}_{s}$, and static dependencies $\mathcal{D}$
\STATE Instantiate the candidate and synthesize example inputs
\STATE Recover FX dependencies $\mathcal{E}_{\mathrm{fx}}$ and
observed contracts through symbolic tracing and shape propagation
\STATE Refine $\mathcal{I}$ and $\mathcal{O}$ with observed contracts
\STATE Generate execution-grounded facts $\mathcal{F}_{e}$
\STATE Set $\mathcal{E}\leftarrow\mathcal{E}_{\mathrm{fx}}$ if
available; otherwise set $\mathcal{E}\leftarrow\mathcal{D}$
\STATE Set $\mathcal{F}\leftarrow
\operatorname{Dedup}(\mathcal{F}_{s}\cup\mathcal{F}_{e})$
\STATE Assemble and render
$G(P)=(\mathcal{I},\mathcal{O},\mathcal{M},
\mathcal{T},\mathcal{F},\mathcal{E})$
\RETURN $G(P)$ and $C(P)$
\end{algorithmic}
\end{algorithm}

The resulting state is serialized using a fixed field order: input
contract, output contract, architecture modules, architecture
operations, structure facts, and architecture graph edges. Executable
FX edges are preferred during rendering; static dependencies are shown
only when executable edges are unavailable. Lookup and indexing nodes
that do not convey useful architectural structure are filtered, and the
number of rendered entities is bounded to keep the mutation context
compact. Node-level shape metadata and profile labels are used
internally but are not emitted as separate fields in the compact
LLM-visible state.
\begin{table*}[t]
\centering
\resizebox{\linewidth}{!}{
\begin{tabular}{lccccccc}
\toprule
Method
& Typed Edge
& Fan-out
& Partition
& Producer
& Next-hop
& Interface/Failure
& Macro Avg. \\
\midrule
Code Only
& 0.5128
& 0.7052
& 0.5769
& 0.3077
& 0.1667
& 0.1667
& 0.4060 \\

CodeRAG-style
& \textbf{0.5385}
& 0.6867
& \textbf{0.6376}
& 0.2308
& 0.2556
& 0.1667
& 0.4193 \\

CPG-style
& 0.5128
& \textbf{0.7697}
& 0.6107
& 0.4615
& 0.3056
& 0.1667
& 0.4712 \\

GraphCode-style
& 0.5128
& 0.7316
& 0.5769
& 0.4615
& 0.2500
& 0.1667
& 0.4499 \\

\method
& \textbf{0.5385}
& 0.7138
& 0.6218
& \textbf{0.5385}
& \textbf{0.3611}
& \textbf{0.2500}
& \textbf{0.5040} \\
\bottomrule
\end{tabular}
}
\caption{Results on NAS-Dependency, constructed from OpenEvolve
evolution traces in the MNIST1D-Shuffle CNN setting. Typed Edge reports
relation accuracy; Fan-out and Next-hop report set F1; Partition reports
macro-F1; Producer and Interface/Failure report exact match. Macro Avg.
is the unweighted mean over the six normalized task metrics.}
\label{tab:nas_dependency}
\end{table*}

\begin{table*}[t]
\centering
\resizebox{\linewidth}{!}{
\begin{tabular}{lccccc}
\toprule
Method
& Model Size $\downarrow$
& DFS Acc. $\uparrow$
& Avg. Acc. $\uparrow$
& Avg. Acc. w/o DFS $\uparrow$
& Wins $\uparrow$ \\
\midrule
EvoPrompting$^{\dagger}$ \cite{evoprompt2023neurips}
& 478.2K
& 68.14
& 80.88
& 81.32
& 2/30 \\
Triplet-GMPNN$^{\dagger}$ \cite{gmpnn2022}
& \textbf{467.3K}
& 47.79
& 75.98
& 76.95
& 1/30 \\
\midrule
OpenEvolve \cite{openevolve}
& 498.6K
& 32.54
& 62.69
& 63.73
& 1/30 \\
Spark \cite{liu2026structured}
& 470.2K
& 83.74
& 83.91
& 83.92
& 5/30 \\
OpenEvolve+\method\ (ours)
& 467.4K
& \textbf{94.68}
& \textbf{89.21}
& \textbf{89.03}
& \textbf{22/30} \\
\bottomrule
\end{tabular}
}
\caption{
CLRS downstream architecture-evolution results.
EvoPrompting$^{\dagger}$ and Triplet-GMPNN$^{\dagger}$, we report the per-task CLRS results from their original papers. 
All other methods are evaluated on CLRS-DFS under Spark's experimental
setting and evaluated zero-shot on all 30 tasks.
Model Size denotes the average number of parameters.
Avg.\ Acc.\ w/o DFS measures transfer to the remaining 29 tasks.
Detailed results on each task can be seen in the supplementary material.
}
\label{tab:clrs_main}
\end{table*}

\subsection{GraphIR-Guided Architecture Mutation}

At iteration $t$, the mutation prompt is constructed from the parent
program $P_t^{\mathrm{par}}$, its GraphIR state $G_t$, and the available
evaluation history $H_t$. The source program provides the exact
implementation to be edited, GraphIR exposes its architecture
components, interface contracts, and producer-consumer relations, and
the evaluation history provides feedback from previously evaluated
candidates.

In the primary configuration, the source program is presented before
the GraphIR context so that the executable implementation remains the
editing anchor. GraphIR is subsequently provided as supplementary
architecture-level evidence. The reverse ordering is evaluated
separately as an ablation.

Mutation remains open-ended and is performed directly in program space.
\method does not impose a predefined architecture grammar or a fixed
set of mutation operators. The LLM may modify module types, hidden
dimensions, branches, tensor transformations, aggregation mechanisms,
residual paths, recurrent structures, or output heads. GraphIR instead
makes the current architecture state explicit, allowing the LLM to
identify meaningful mutation targets and reason about the interfaces
and downstream consumers affected by a proposed edit.

After the child program is generated, the underlying evolution
framework validates and evaluates it and updates the candidate archive.
GraphIR is then reconstructed from the child program, so newly
introduced modules, operations, contracts, structure facts, and
dataflow dependencies become available in subsequent mutation
iterations.

\begin{table*}[t]
\small
\centering
\small
\begin{tabular}{lcccccc}
\toprule
Backbone
& Seeds
& $n$
& Acc. $\uparrow$
& Model Size $\downarrow$
& Invalid $\downarrow$
& GPU-hours $\downarrow$ \\
\midrule
CNN
& 42--51
& 10
& \textbf{71.55 $\pm$ 2.14}
& 28.74K $\pm$ 17.84K
& 19.80
& 1.639 $\pm$ 0.089 \\

MLP
& 42--47
& 6
& 70.42 $\pm$ 0.79
& 23.70K $\pm$ 5.08K
& \textbf{4.50}
& 1.048 $\pm$ 0.049 \\

GRU
& 42--47
& 6
& 66.18 $\pm$ 3.32
& 7.35K $\pm$ 13.99K
& 9.50
& 1.557 $\pm$ 0.076 \\

Linear
& 42--47
& 6
& 56.38 $\pm$ 1.66
& \textbf{1.40K $\pm$ 0.43K}
& 4.83
& \textbf{0.829 $\pm$ 0.089} \\
\bottomrule
\end{tabular}
\caption{Robustness across backbone-specific search spaces and random
seeds on MNIST1D-Shuffle. Results report mean $\pm$ standard deviation
over independent architecture-evolution runs.
Qwen3-Plus
~\cite{yang2025qwen3} is used as the evaluation LLM.}
\label{tab:seed_backbone}
\end{table*}
\begin{table*}[t]
\centering
\small
\begin{tabular}{lccccc}
\toprule
\textbf{Prompt Order}
& \textbf{Acc. $\uparrow$}
& \textbf{Size $\downarrow$}
& \textbf{Invalid $\downarrow$}
& \textbf{Shape Err. $\downarrow$}
& \textbf{GPU-hours $\downarrow$} \\
\midrule
Code $\rightarrow$ \method
& \textbf{74.20}
& \textbf{26.49K}
& 21
& \textbf{6}
& \textbf{1.605} \\
\method $\rightarrow$ Code
& 66.30
& 35.55K
& \textbf{18}
& 17
& 2.326 \\
\bottomrule
\end{tabular}
\caption{Effect of source--state ordering in MNIST1D-Shuffle CNN setting. Results
report the best observed run for each ordering. NAS time denotes the
end-to-end wall-clock time of the complete search pipeline on a system
equipped with a single NVIDIA RTX 3090 GPU.}
\label{tab:prompt_order}
\end{table*}

\begin{table*}[t]
\centering
\resizebox{\linewidth}{!}{
\begin{tabular}{lccccccc}
\toprule
Variant
& Node Facts
& Edge Relations
& Max Test Acc. $\uparrow$
& Model Size $\downarrow$
& Invalid $\downarrow$
& Shape Mismatch $\downarrow$
& Runtime (GPU-h) $\downarrow$ \\
\midrule

\method{}
& $\checkmark$
& $\checkmark$
& $\mathbf{73.53 \pm 0.83}$
& $30.86 \pm 8.06$K
& $19.67 \pm 7.09$
& $\mathbf{7.33 \pm 5.13}$
& $\mathbf{1.629 \pm 0.098}$ \\

\method{} w/o edge relations
& $\checkmark$
& $\times$
& $71.53 \pm 0.40$
& $23.61 \pm 16.41$K
& $\mathbf{14.67 \pm 7.77}$
& $14.33 \pm 8.33$
& $2.034 \pm 0.287$ \\

\method{} w/o node facts
& $\times$
& $\checkmark$
& $67.13 \pm 3.20$
& $22.18 \pm 13.37$K
& $16.67 \pm 6.51$
& $16.67 \pm 6.51$
& $2.216 \pm 0.223$ \\

\bottomrule
\end{tabular}
}
\caption{Ablation of the information retained in the GraphIR state in the
MNIST1D-Shuffle CNN setting. Results are reported as mean $\pm$ standard deviation
over three random seeds.}
\label{tab:state_content}
\end{table*}

\section{Experiments}

We organize our experiments around three research questions that examine
the reasoning mechanism, downstream effectiveness, and design choices of
\method:

\begin{itemize}
    \item \textbf{RQ1: What architecture-level information can LLMs
    effectively utilize from \method?}
    We evaluate whether \method helps an LLM recover the typed
    dependencies, producer occurrences, propagation paths, and interface
    risks required for architecture mutation.

    \item \textbf{RQ2: Does \method improve LLM-guided neural
    architecture evolution?}
    We compare \method with existing LLM-guided search methods and
    alternative structured representations across diverse architecture
    search tasks, considering search performance, model size, and
    end-to-end NAS time.

    \item \textbf{RQ3: Which components and usage choices of \method
    are responsible for the improvement?}
    We examine robustness across random seeds and architecture families,
    and ablate the source--state ordering, node facts, and edge relations
    used in the rendered architecture state.
\end{itemize}

RQ1 isolates whether the representation exposes useful architecture
information, RQ2 tests whether this reasoning advantage translates into
better search outcomes, and RQ3 identifies the design choices underlying
the observed improvements.

\subsection{Experimental Setup}

All architecture-evolution experiments are implemented within the
OpenEvolve framework \cite{openevolve}, with Qwen3-Plus used to generate candidate
programs. Unless otherwise specified, compared configurations share the
same search and evaluation procedures and differ only in the mutation
context supplied to the LLM. Since the three research questions use
different tasks and metrics, their specific settings are introduced in
the corresponding subsections.
\paragraph{Evaluation metrics.}
We report test accuracy (\emph{Acc.}) on the standard test set and
out-of-distribution accuracy (\emph{OOD Acc.}) on the corresponding
distribution-shifted test set.
\emph{Model Size} denotes the total number of parameters in the
evaluated candidate architecture, reported in thousands.
\emph{Invalid}
counts generated candidates that cannot complete execution or
evaluation, while \emph{Shape Err.} counts candidates that fail because
of incompatible tensor shapes. \emph{Stage-2 Eval.} denotes the number
of candidates admitted to full evaluation after preliminary validation.
\emph{GPU-hours} reports the end-to-end wall-clock time of the complete
search pipeline, including candidate generation, program analysis,
validation, and candidate evaluation, on a system equipped with a
single NVIDIA RTX 3090 GPU. Unless otherwise specified, multi-run
results are reported as mean and standard deviation.

\subsection{RQ1: Architecture-Level Reasoning}

To examine whether existing structured code contexts expose the
architecture dependencies required for NAS editing, we construct
\emph{NAS-Dependency}, a benchmark generated from evolved CNN programs
in the MNIST1D-Shuffle NAS setting. It contains 120 questions covering
six complementary dimensions: typed-edge recognition tests relation
typing, value fan-out recovery tests downstream coverage, dependency
partitioning tests structural grouping, producer-occurrence
identification tests exact producer localization, dependency next-hop
prediction tests local dependency propagation, and interface or failure
diagnosis tests compatibility-risk identification. 

We compare the code-only OpenEvolve context with three generic code
representations adapted to NAS, namely \textsc{CodeRAG-style} \cite{du2024codegrag},
\textsc{CPG-style} \cite{yamaguchi2014modeling}, and \textsc{GraphCode-style} \cite{liu2024graphcoder} , as well as
\method. We use the suffix ``-style'' because these baselines
adapt the representation principles of the original methods to NAS
candidate programs rather than reproducing their complete systems.
The compared settings differ only in the additional context provided
to the LLM.

As shown in Table~\ref{tab:nas_dependency}, \method achieves the highest
macro average and the best or tied-best result on four of the six
dimensions. Its advantages concentrate on producer-occurrence
identification, dependency next-hop prediction, and interface or failure
diagnosis, which respectively require locating the exact producer,
tracing downstream dependency propagation, and identifying potentially
incompatible interfaces. These results suggest that \method provides
more than a generic code graph: it exposes an occurrence-aware and
interface-aware architecture state tailored to mutation. CPG-style
remains stronger on fan-out recovery, while CodeRAG-style performs best
on dependency partitioning, indicating that generic structured contexts
remain competitive for local structural reasoning. Detailed task
definitions and representative cases are provided in the supplementary
material.

\subsection{RQ2: Architecture Evolution Performance}

\paragraph{Results on CLRS.}
We evaluate whether the architecture-level reasoning improvements
observed on NAS-Dependency translate into improved architecture
evolution performance on CLRS~\cite{clrs2024}, a benchmark comprising
30 algorithmic reasoning tasks.
We compare against SPARK~\cite{liu2026structured}, EvoPrompting~\cite{evoprompt2023neurips}, OpenEvolve~\cite{openevolve}, and Triplet-GMPNN~\cite{gmpnn2022}.
As shown in Table~\ref{tab:clrs_main}, OpenEvolve+\method achieves the
best DFS accuracy and the highest average accuracy across all 30 CLRS
tasks, while maintaining a model size comparable to the original
Triplet-GMPNN. Excluding the DFS search task, it reaches an average
accuracy of $89.03\%$ and obtains the best result on 22 of the 30 tasks,
demonstrating that the architectures discovered with \method transfer
effectively beyond the task used for evolution. 

\begin{table*}[t]
\centering
\small
\begin{tabular}{lccccccc}
\toprule
\textbf{Method}
& \textbf{Runs}
& \textbf{Acc. $\uparrow$}
& \textbf{OOD Acc. $\uparrow$}
& \textbf{Params $\downarrow$}
& \textbf{Invalid $\downarrow$}
& \textbf{Shape Err. $\downarrow$}
& \textbf{GPU-hours (h) $\downarrow$} \\
\midrule
Initial
& -- & 54.70 & -- & 5.21K & -- & -- & -- \\
OpenEvolve
& 1 & 59.10 & -- & 5.36K & -- & -- & 1.270 \\
\midrule
CodeGRAG-style
& 3
& $66.40 \pm 3.57$
& $64.87 \pm 2.75$
& $\mathbf{15.27 \pm 9.70}$K
& $20.33$
& $7.67$
& $\mathbf{1.621 \pm 0.153}$ \\
CPG-style
& 3
& $70.10 \pm 3.47$
& $67.95 \pm 3.12$
& $21.52 \pm 3.84$K
& $22.67$
& $11.33$
& $2.124 \pm 0.184$ \\
GraphCode-style
& 3
& $68.03 \pm 1.91$
& $66.20 \pm 1.02$
& $18.34 \pm 8.93$K
& $\mathbf{15.67}$
& $\mathbf{4.67}$
& $2.411 \pm 0.252$ \\
\method
& 10
& $\mathbf{71.55 \pm 2.14}$
& $\mathbf{68.68 \pm 2.53}$
& $28.74 \pm 17.84$K
& $19.80$
& $6.80$
& $1.639 \pm 0.089$ \\
\bottomrule
\end{tabular}
\caption{Comparison with alternative structured representations in the
MNIST1D-Shuffle CNN setting. All methods use the same OpenEvolve
pipeline, Qwen3-Plus mutation model, prompt template, evaluation
protocol, and budget of 100 evolution iterations; only the structured
representation supplied to the LLM is replaced. Results for the
structured representations report mean and standard deviation over
three seeds for CodeGRAG-style, CPG-style, and GraphCode-style, and ten
seeds for \method. NAS time denotes the end-to-end wall-clock time of
the complete search pipeline on a system equipped with a single
NVIDIA RTX 3090 GPU. Bold indicates the best result among the
structured representations. }
\label{tab:representation_comparison}
\end{table*}
\paragraph{Comparison with alternative structured representations.}
To isolate the effect of the architecture representation, we retain the
same OpenEvolve pipeline, mutation prompt, LLM, search budget, and
evaluation protocol, and replace only the \method state with
CodeGRAG-style, CPG-style, or GraphCode-style representations. As shown
in Table~\ref{tab:representation_comparison}, \method achieves the
highest average in-distribution and OOD accuracies, reaching
$71.55\%$ and $68.68\%$, respectively. It also requires substantially
less end-to-end NAS time than CPG-style and GraphCode-style. These
results indicate that the gains of \method do not arise merely from
providing an additional graph, but from exposing a mutation-aligned
architecture state tailored to neural architecture evolution.

\subsection{RQ3: Robustness and Ablation Studies}

We first examine the robustness of \method across random seeds and
network backbones. We then study two design choices: the ordering of
the source program and architecture state, and the information retained
in the state.

\paragraph{Robustness across search spaces and random seeds.}
We evaluate the robustness of \method to both search stochasticity and
architectural heterogeneity through independent evolution runs across
multiple random seeds and four backbone-specific search spaces on
MNIST1D-Shuffle: CNN, MLP, GRU, and linear models.
As shown in
Table~\ref{tab:seed_backbone}, \method maintains relatively consistent
performance across repeated runs for CNN, MLP, GRU, and linear
architectures.

\paragraph{Source--state ordering.}
We compare a code-first prompt, where the source program precedes
\method, with the reverse state-first ordering. The code-first
configuration achieves a substantially higher best test accuracy
($74.20\%$ vs.\ $66.30\%$), while producing fewer shape mismatches
($6$ vs.\ $17$) and requiring less end-to-end NAS GPU time
($1.605$ vs.\ $2.326$ hours). Although the state-first ordering yields
slightly fewer invalid candidates, its lower accuracy and higher shape
mismatch rate indicate that presenting the architecture state before the
executable program may weaken the code context needed for reliable
editing. These results support using the source program as the primary
editing anchor, with \method serving as supplementary architecture-level
guidance.

\paragraph{Ablation on architecture-state content.}
We separately ablate two components of the GraphIR state. \emph{Node
facts} describe architecture entities and their local attributes,
including module or tensor-operation types and available shape or
interface information. \emph{Edge relations} encode the
producer-consumer and dependency links among these entities. As shown
in Table~\ref{tab:state_content}, removing either component degrades
search performance and increases shape mismatches. The larger drop
after removing node facts suggests that explicit architecture semantics
are particularly important for identifying meaningful mutation targets,
while edge relations support dependency-aware and structurally
compatible edits. Retaining both components yields the highest accuracy
and the lowest runtime.

\section{Conclusion}

We introduced \method, a mutation-aligned architecture representation for
LLM-guided code-based NAS. Instead of relying on source code alone,
\method makes the candidate architecture explicit through its computation
skeleton, mutation surface, and validity envelope, while preserving the
program as the editable search object.
Experiments on \emph{NAS-Dependency} and six downstream
architecture-search benchmarks show that \method improves architecture
reasoning, search quality, and end-to-end efficiency across diverse
settings. In particular, it helps LLMs localize producers, trace
dependency propagation, and identify interface risks that are difficult
to recover from implementation details alone. These findings suggest
that the representation of candidate architectures is a central design
choice in LLM-guided NAS, and that mutation-aligned states provide a
practical interface between executable neural programs and architecture
evolution.

\bibliography{aaai2027}


\providecommand{\method}{GraphIR}
\providecommand{\code}[1]{\texttt{#1}}

\definecolor{graphirlistingbg}{gray}{0.955}
\lstdefinestyle{graphirqa}{
    basicstyle=\ttfamily\scriptsize,
    numbers=none,
    breaklines=true,
    breakatwhitespace=false,
    breakindent=0pt,
    columns=fullflexible,
    keepspaces=true,
    frame=none,
    backgroundcolor=\color{graphirlistingbg},
    xleftmargin=0.55em,
    xrightmargin=0.55em,
    aboveskip=0.35em,
    belowskip=0.35em,
    showstringspaces=false,
    tabsize=2
}
\lstset{style=graphirqa}
\renewcommand{\arraystretch}{1.10}
\newcommand{\goldanswer}{\noindent\textit{Gold answer.}\par}
\newcommand{\qaheading}[1]{\par\medskip\noindent\textbf{#1.}\par\smallskip}

\appendix
\setcounter{secnumdepth}{1}

\section*{Supplementary Material}
\noindent
This supplement provides the complete construction and evaluation details
for NAS-Dependency, representative benchmark instances, downstream
experimental settings, and full numerical results. Sections~A--C describe
the benchmark and its evaluation protocol, while Sections~D--F report the
downstream and per-task results.

\section{NAS-Dependency Benchmark}
\label{sec:nas_dependency}

\subsection{Purpose and Separation from GraphIR}

\emph{NAS-Dependency} evaluates whether an LLM can recover the
architecture-level dependencies required for neural architecture
mutation. The benchmark is generated from real parent--child records
collected during code-based NAS.

Two components must be clearly distinguished. \emph{SourceOracleV2} is
the deterministic source analyzer used only to construct and verify the
benchmark. It extracts normalized structural facts from the child
program. 
GraphIR is the representation evaluated by the paper. It is
provided to the LLM as an optional additional representation and does
not participate in question construction or Gold-answer generation.

\subsection{Source Records}

Each source record corresponds to one parent-to-child architecture
transition and stores:
\begin{itemize}
    \item the parent program;
    \item the child program;
    \item the parent-to-child source diff;
    \item search and evaluation metadata, including the experiment,
    variant, prompt order, random seed, iteration, source-record
    identifier, and execution status.
\end{itemize}

The child program is the primary source for structural questions. The
parent program and source diff are retained for provenance but are not
provided to the evaluated LLM. Runtime metadata supplies independently
observed shape and failure evidence for interface and failure questions.

\subsection{Construction Pipeline}

The construction process is organized into four deterministic stages,
summarized in Table~\ref{tab:construction_pipeline}.

\begin{center}
\small
\setlength{\tabcolsep}{4pt}
\begin{minipage}{0.98\columnwidth}
\centering
\begin{tabular}{cp{0.77\columnwidth}}
\toprule
\textbf{Stage} & \textbf{Operation} \\
\midrule
1 & Read a parent--child NAS record, the child code, and independent evaluation evidence. \\
2 & Parse the child with SourceOracleV2 and construct canonical nodes and typed semantic edges. \\
3 & Extract unambiguous facts, generate candidate sets and validated distractors, and instantiate fixed templates. \\
4 & Compute and independently verify Gold answers, then split by source record into Dev-44 and Held-out-76. \\
\bottomrule
\end{tabular}
\captionof{table}{Deterministic construction pipeline for NAS-Dependency. The final dataset contains six task families and 120 QA instances.}
\label{tab:construction_pipeline}
\end{minipage}
\end{center}

Question text, candidate sets, and Gold answers are generated from the
same verified fact. No LLM is used to generate questions or standard
answers.

\subsection{Canonical Source Graph}

SourceOracleV2 first parses the child program with the Python abstract
syntax tree. For parsable programs, it locates the model's
\code{forward} scope and constructs a normalized graph
\(G=(V,E)\).

\paragraph{Canonical nodes.}
The graph contains:
\begin{itemize}
    \item tensor versions, such as \code{h3\#2};
    \item operation occurrences, such as \code{call.view@2};
    \item module occurrences, such as \code{self.conv3@1};
    \item metadata operations, such as \code{shape\_index@3};
    \item unary operations, such as \code{usub@5};
    \item control-flow merge nodes, such as \code{phi:y@1}; and
    \item a unified \code{output} node.
\end{itemize}

Tensor-version counters distinguish repeated assignments to the same
Python variable. Operation-occurrence counters distinguish repeated
source-level occurrences of the same operation. Occurrence indices
refer to source order, not dynamic execution count.

\paragraph{Typed edges.}
The graph uses the following semantic relations:
\begin{table*}[t]
\centering
\small
\begin{tabular}{lp{0.78\textwidth}}
\toprule
\textbf{Edge type} & \textbf{Meaning} \\
\midrule
\code{READ\_VALUE} & The target operation directly reads the numerical value of the source tensor. \\
\code{READ\_METADATA} & The target operation reads only metadata such as shape, device, or dtype. \\
\code{PRODUCES} & The source operation directly produces the target tensor version. \\
\code{RETURNS} & The source value is returned by \code{forward}. \\
\code{ALIAS\_OF} & Two nodes refer to the same underlying value. \\
\code{PHI\_INPUT} & A branch value enters a control-flow merge node. \\
\code{CONTROLS} & A condition or loop controls a statement or operation. \\
\bottomrule
\end{tabular}
\caption{Semantic edge types used by SourceOracleV2.}
\label{tab:oracle_edges}
\end{table*}

For example, the statement
\code{y = h.view(h.shape[0], -1)} yields a direct
\code{READ\_VALUE} edge from the current version of \code{h} to the
outer \code{view} occurrence, a \code{READ\_METADATA} edge from
\code{h} to the shape-index occurrence, and a \code{PRODUCES} edge from
the outer \code{view} occurrence to the new tensor version of \code{y}.

If AST parsing fails, SourceOracleV2 does not construct dependency-graph
questions. It retains only independently verifiable failure facts, such
as \code{syntax\_error}, which may be used for Interface/Failure
questions.

\subsection{Fact Extraction}

The normalized graph is converted into task-specific facts:
\begin{itemize}
    \item \(\mathrm{DirectEdge}(u,v)\);
    \item \(\mathrm{ValueConsumers}(x)\);
    \item \(\mathrm{MetadataConsumers}(x)\);
    \item \(\mathrm{Producer}(x)\);
    \item \(\mathrm{ShortestPathNextHop}(u,\mathrm{output})\);
    \item observed module input and output shapes; and
    \item deterministic failure evidence.
\end{itemize}

A fact is retained only when its nodes map back to source code, its
relation is unambiguous, its candidate set is valid, and its Gold answer
is unique or set-valued under a deterministic definition. Distractors
are sampled from structurally plausible but oracle-verified incorrect
nodes, including wrong operation occurrences, metadata-only consumers,
transitive descendants, nested operand calls, and nearby nodes on other
branches.

\subsection{Question and Gold Generation}

Each task family uses a fixed template. The selected fact fills the
query target and correct candidate, while validated distractors fill the
remaining candidate set. The Gold JSON is computed directly from the
fact. A task-specific response contract then specifies the required JSON
schema, allowed identifiers, and output restrictions.

After construction, a validator rereads the SourceOracleV2 graph and
recomputes the Gold answer. All 120 retained QA instances pass this
oracle-consistency check. This guarantees agreement between the Gold
answer and the normalized oracle graph. It does not imply that
SourceOracleV2 is a complete interpreter for arbitrary dynamic Python.

\subsection{Dataset Split and Statistics}

The split is performed by source record rather than by individual
question. All questions generated from the same child program remain in
the same split.

\begin{center}
\small
\begin{minipage}{0.92\columnwidth}
\centering
\begin{tabular}{lrr}
\toprule
\textbf{Split} & \textbf{Source records} & \textbf{QA instances} \\
\midrule
Development & 6 & 44 \\
Held-out & 14 & 76 \\
\midrule
Total & 20 & 120 \\
\bottomrule
\end{tabular}
\captionof{table}{NAS-Dependency split statistics. All reported benchmark results
use the held-out split.}
\label{tab:nas_dependency_split}
\end{minipage}
\end{center}

The benchmark contains six task families with 20 questions per family:
\begin{center}
\small
\begin{minipage}{0.92\columnwidth}
\centering
\begin{tabular}{lr}
\toprule
\textbf{Task family} & \textbf{Number of questions} \\
\midrule
Typed Edge & 20 \\
Value Fan-out & 20 \\
Partition & 20 \\
Producer Occurrence & 20 \\
Dependency Next-hop & 20 \\
Interface/Failure & 20 \\
\midrule
Total & 120 \\
\bottomrule
\end{tabular}
\captionof{table}{Task distribution in NAS-Dependency.}
\label{tab:nas_dependency_distribution}
\end{minipage}
\end{center}

\subsection{Task Definitions and Metrics}

\begin{table*}[t]
\centering
\small
\begin{tabular}{p{0.17\textwidth}p{0.43\textwidth}p{0.29\textwidth}}
\toprule
\textbf{Task} & \textbf{Required reasoning} & \textbf{Primary metric} \\
\midrule
Typed Edge
& Classify the direct semantic relation for every candidate node pair.
& Relation accuracy \\
Value Fan-out
& Select all candidate operations that directly consume a queried tensor value.
& Set F1 \\
Partition
& Partition candidates into value consumers, metadata consumers, and non-consumers.
& Partition macro-F1 \\
Producer Occurrence
& Identify the exact outermost operation occurrence that directly produces a tensor version.
& Exact match \\
Dependency Next-hop
& Select all candidate nodes that can immediately follow the start node on a shortest valid path to \code{output}.
& Set F1 \\
Interface/Failure
& Infer an observed output shape, affected interface, or deterministic failure category.
& Exact match \\
\bottomrule
\end{tabular}
\caption{The six NAS-Dependency task families.}
\label{tab:nas_dependency_tasks}
\end{table*}

Each task also reports strict exact match, which equals one only when the
entire QA instance is answered correctly.

\FloatBarrier

\section{Representative Benchmark Examples}
\label{sec:qa_examples}

This section gives one representative instance for each task family. To improve readability, each block contains only the task query, candidate set, and deterministic Gold JSON; the complete 120-instance benchmark is distributed separately as structured data.

\qaheading{Typed Edge}

\begin{lstlisting}
Classify the direct edge type for each candidate pair below.
Use `NONE` when no direct edge exists.

Candidate pairs:
[
  {"candidate_id": "R1", "source": "h2#5", "target": "add@3"},
  {"candidate_id": "R2", "source": "h2#5",
   "target": "self.temporal_attn_out@1"},
  {"candidate_id": "R3", "source": "add@3", "target": "h2#6"}
].

Return one classification for every candidate_id.
\end{lstlisting}

\goldanswer
\begin{lstlisting}
{
  "relations": [
    {"candidate_id": "R1", "edge_type": "READ_VALUE"},
    {"candidate_id": "R2", "edge_type": "NONE"},
    {"candidate_id": "R3", "edge_type": "PRODUCES"}
  ]
}
\end{lstlisting}

\qaheading{Value Fan-out}

\begin{lstlisting}
Select every canonical operation occurrence from the candidate list
that directly consumes the tensor value of `h3_scaled#1`.

Exclude metadata-only edges and transitive downstream consumers.

Candidates:
[
  "self.residual_pool@1",
  "self.residual_proj@1",
  "call.mean@1",
  "call.permute@1",
  "mul@4",
  "call.view@1",
  "mul@3",
  "call.view@2",
  "shape_index@2",
  "call.view@3"
].
\end{lstlisting}

\goldanswer
\begin{lstlisting}
{
  "selected_operations": [
    "call.mean@1",
    "call.permute@1",
    "call.view@2",
    "mul@4",
    "self.residual_proj@1"
  ]
}
\end{lstlisting}

\qaheading{Partition}

\begin{lstlisting}
For `h2#1`, partition every listed canonical operation into exactly
one class: direct value consumer, direct metadata consumer, or not
a direct consumer.

Candidates:
[
  "call.view@2",
  "self.attention_pool@1",
  "self.attention_pool_max@1",
  "shape_index@2",
  "usub@2",
  "shape_index@5",
  "shape_index@4",
  "shape_index@3",
  "shape_index@6",
  "mul@1"
].
\end{lstlisting}

\goldanswer
\begin{lstlisting}
{
  "value_consumers": [
    "mul@1",
    "self.attention_pool@1",
    "self.attention_pool_max@1"
  ],
  "metadata_consumers": [
    "shape_index@2",
    "shape_index@3",
    "shape_index@4",
    "shape_index@5",
    "shape_index@6"
  ],
  "non_consumers": [
    "call.view@2",
    "usub@2"
  ]
}
\end{lstlisting}

\qaheading{Producer Occurrence}

\begin{lstlisting}
Which listed outermost operation node has a direct `PRODUCES` edge
to `attn_scores_flat#1`?

Choose the operation whose result is bound to the tensor; do not
choose nested calls used only as operands.

Candidates:
[
  "torch.matmul@1",
  "usub@5",
  "mul@3",
  "call.transpose@1",
  "usub@4",
  "mul@5"
].
\end{lstlisting}

\goldanswer
\begin{lstlisting}
{
  "producer": "mul@5"
}
\end{lstlisting}

\qaheading{Dependency Next-hop}

\begin{lstlisting}
Consider all shortest directed dependency paths from `h3#2` to
`output` using the allowed semantic edge kinds.

Select the canonical graph node or nodes that can appear immediately
after the start node on such a shortest path.

Return only identifiers from this candidate list:
[
  "call.sum@1",
  "flat_h3#1",
  "call.view@2",
  "gap_h3#1",
  "self.temporal_attention@1",
  "mul@1",
  "call.mean@1"
].
\end{lstlisting}

\goldanswer
\begin{lstlisting}
{
  "next_nodes": [
    "call.mean@1"
  ]
}
\end{lstlisting}

\qaheading{Interface/Failure}

\noindent\textit{Observed-shape example.}\par
\begin{lstlisting}
The observed input shape to `self.conv2@1` is ["B", 25, 19].
What is its observed output shape under the same execution?
\end{lstlisting}

\goldanswer
\begin{lstlisting}
{
  "output_shape": [
    "B",
    25,
    10
  ]
}
\end{lstlisting}

\noindent\textit{Failure-diagnosis variant.}\par
\begin{lstlisting}
The child architecture is independently known to be invalid.
What is the earliest deterministic failure category observed
for this program?
\end{lstlisting}

\goldanswer
\begin{lstlisting}
{
  "failure_type": "elementwise_shape_mismatch"
}
\end{lstlisting}

\FloatBarrier

\section{Evaluation Protocol}
\label{sec:dependency_protocol}

\subsection{Compared Input Representations}

For every held-out QA, all methods receive the same system message,
child code, question, response contract, LLM, generation parameters,
parser, and deterministic scorer. The only controlled difference is the
additional representation:

\begin{table*}[t]
\centering
\small
\begin{tabular}{lp{0.78\textwidth}}
\toprule
\textbf{Method} & \textbf{Input supplied to the LLM} \\
\midrule
Code Only & Child Code + Question + Response Contract \\
CodeRAG-style & Child Code + retrieved source snippets + Question + Response Contract \\
CPG-style & Child Code + statement nodes and ordering relations + Question + Response Contract \\
GraphCode-style & Child Code + statement-level def-use relations + Question + Response Contract \\
\method & Child Code + compact architecture-aware state + Question + Response Contract \\
\bottomrule
\end{tabular}
\caption{Controlled input conditions for NAS-Dependency.}
\label{tab:dependency_inputs}
\end{table*}

The evaluated LLM never receives the parent program, source diff, Gold
answer, or complete SourceOracleV2 graph.

\subsection{Unified Prompt}

The exact prompt template used in every condition is shown below.

\begin{lstlisting}
[SYSTEM MESSAGE]

You are evaluating dependency reasoning over an executable neural
network child program.

Use only the supplied Child Code, optional Additional Representation,
Question, and Response Contract.

The Question contains the complete query and candidate set. Identifiers
such as `{variable_name}#{version}` and
`{operation_label}@{occurrence}` refer to specific variable versions
and source-level operation occurrences in the Child Code.

Carefully distinguish:
- direct dependencies from transitive dependencies;
- tensor-value reads from metadata-only reads;
- outermost operations from nested operand-only operations; and
- one operation occurrence from other occurrences with the same label.

Follow the Response Contract exactly.
Return only one valid JSON object.

[USER MESSAGE]

[Child Code]
{child_code}

[Additional Representation: {method_name}]
{additional_representation}

[Question]
{question}

[Response Contract]
{response_contract}
\end{lstlisting}

The Additional Representation block is omitted for Code Only.

\subsection{Model and Generation Settings}

All conditions use Qwen3-Plus through the \code{qwen-plus} API alias
with temperature \(0.0\), maximum output length 1536 tokens, timeout
120 seconds, and at most two retries for API failures. Each
(QA, method) pair produces one final prediction. The complete evaluation
contains \(120\times5=600\) predictions, while paper results are reported
on the 76 held-out questions.

\subsection{Deterministic Scoring}

Predictions are parsed as JSON and validated against the task-specific
response contract. Scoring does not use an LLM judge. The paper reports
per-task scores, task macro average, source macro average, micro primary
score, and strict exact match.

\subsection{Held-out Results}

\begin{table*}[t]
\centering
\small
\begin{tabular}{lrrrrrrr}
\toprule
\textbf{Method}
& \textbf{Edge}
& \textbf{Fan-out}
& \textbf{Partition}
& \textbf{Producer}
& \textbf{Next-hop}
& \textbf{Interface}
& \textbf{Macro} \\
\midrule
Code Only
& .5128 & .7052 & .5769 & .3077 & .1667 & .1667 & .4060 \\
CodeRAG-style
& \textbf{.5385} & .6867 & \textbf{.6376} & .2308 & .2556 & .1667 & .4193 \\
CPG-style
& .5128 & \textbf{.7697} & .6107 & .4615 & .3056 & .1667 & .4712 \\
GraphCode-style
& .5128 & .7316 & .5769 & .4615 & .2500 & .1667 & .4499 \\
\method
& \textbf{.5385} & .7138 & .6218 & \textbf{.5385}
& \textbf{.3611} & \textbf{.2500} & \textbf{.5040} \\
\bottomrule
\end{tabular}
\caption{Held-out results on NAS-Dependency. Edge reports relation
accuracy; Fan-out and Next-hop report set F1; Partition reports
macro-F1; Producer and Interface report exact match. Macro is the
unweighted mean of the six normalized task metrics.}
\label{tab:dependency_results}
\end{table*}

\section{Downstream Benchmark Details}
\label{sec:downstream_details}

\subsection{Benchmarks and Search Initialization}

The downstream evaluation covers six architecture-search benchmarks:
CLRS for algorithmic reasoning, MNIST1D-Shuffle for one-dimensional
sequence classification, 20 Newsgroups for text classification,
CartPole-v1 for reinforcement-learning control, and Yeast and Breast
Cancer Wisconsin Diagnostic for scientific tabular classification.

MNIST1D-Shuffle includes controlled searches over linear, MLP, CNN, and
GRU architecture families. CLRS starts from the benchmark-defined
initial architecture. The remaining benchmarks use random-start
evolution.

\subsection{Evaluation Metrics}

\emph{Accuracy} denotes the task-specific test metric. For
MNIST1D-Shuffle, \emph{OOD accuracy} is measured on the corresponding
distribution-shifted test set. \emph{Model Size} denotes the total number
of parameters in the evaluated candidate architecture.
\emph{Invalid} counts generated candidates that cannot complete
execution or evaluation. \emph{Shape Error} counts candidates whose
failure is attributed to incompatible tensor shapes.
\emph{Stage-2 Evaluation} denotes candidates admitted to full evaluation
after preliminary validation. \emph{NAS GPU hours} is the end-to-end
wall-clock time of the complete search pipeline, including candidate
generation, program analysis, validation, and candidate evaluation, on a
system equipped with one NVIDIA RTX 3090 GPU.

\section{MNIST1D-Shuffle Results}
\label{sec:mnist_results}

All search methods use 100 evolution iterations. The average column is
the arithmetic mean over linear, MLP, CNN, and GRU architecture
families.

\begin{table}[t]
\centering
\small
\setlength{\tabcolsep}{5pt}
\textbf{(a) Test accuracy (\%)}\\[0.25em]
\begin{tabular}{lrrrrr}
\toprule
\textbf{Method} & \textbf{Linear} & \textbf{MLP} & \textbf{CNN} & \textbf{GRU} & \textbf{Average} \\
\midrule
Initial & 31.40 & 65.60 & 54.70 & 50.50 & 50.55 \\
EOH & 57.60 & 65.50 & 59.40 & 67.80 & 62.58 \\
FunSearch & 61.10 & 65.60 & 60.10 & 55.30 & 60.53 \\
EvoPrompting & \textbf{69.80} & 69.90 & 58.70 & 60.60 & 64.75 \\
OpenEvolve & 60.60 & 68.00 & 59.10 & 58.90 & 61.65 \\
\method & 59.30 & \textbf{71.90} & \textbf{74.20} & \textbf{70.80} & \textbf{69.05} \\
\bottomrule
\end{tabular}

\medskip
\textbf{(b) End-to-end NAS time (GPU-hours)}\\[0.25em]
\begin{tabular}{lrrrrr}
\toprule
\textbf{Method} & \textbf{Linear} & \textbf{MLP} & \textbf{CNN} & \textbf{GRU} & \textbf{Average} \\
\midrule
EOH & 5.450 & 1.660 & 2.170 & 39.860 & 12.285 \\
FunSearch & \textbf{0.630} & 1.610 & 2.810 & 13.600 & 4.663 \\
EvoPrompting & 0.720 & 0.740 & \textbf{1.110} & \textbf{0.440} & \textbf{0.753} \\
OpenEvolve & 1.640 & \textbf{0.590} & 1.270 & 0.600 & 1.025 \\
\method & 0.754 & 1.012 & 1.605 & 1.550 & 1.230 \\
\bottomrule
\end{tabular}

\medskip
\textbf{(c) Model size (K parameters)}\\[0.25em]
\begin{tabular}{lrrrrr}
\toprule
\textbf{Method} & \textbf{Linear} & \textbf{MLP} & \textbf{CNN} & \textbf{GRU} & \textbf{Average} \\
\midrule
Initial & 0.410 & 15.210 & 5.210 & 5.134 & 6.491 \\
EOH & 17.682 & 15.210 & 11.050 & 11.002 & 13.736 \\
FunSearch & \textbf{0.410} & 15.210 & 8.202 & \textbf{5.134} & 7.239 \\
EvoPrompting & 23.562 & 31.306 & 6.610 & 14.538 & 19.004 \\
OpenEvolve & 7.460 & 18.826 & \textbf{5.360} & 42.250 & 18.474 \\
\method & 2.061 & 25.510 & 26.485 & 35.890 & 22.487 \\
\bottomrule
\end{tabular}
\caption{Complete MNIST1D-Shuffle results over linear, MLP, CNN, and GRU search spaces. All searches use 100 evolution iterations. Boldface indicates the best value under the criterion used in each panel.}
\label{tab:mnist_complete}
\end{table}

\section{Cross-Task Results}
\label{sec:cross_task}

Table~\ref{tab:cross_task_raw} reports the raw score and runtime values used for aggregation.

\begin{table*}[t]
\centering
\small
\begin{tabular}{lrrrrrrrr}
\toprule
& \multicolumn{2}{c}{\textbf{20News}}
& \multicolumn{2}{c}{\textbf{CartPole-v1}}
& \multicolumn{2}{c}{\textbf{Yeast}}
& \multicolumn{2}{c}{\textbf{Breast Cancer}} \\
\cmidrule(lr){2-3}
\cmidrule(lr){4-5}
\cmidrule(lr){6-7}
\cmidrule(lr){8-9}
\textbf{Method}
& \textbf{Score} & \textbf{Time}
& \textbf{Score} & \textbf{Time}
& \textbf{Score} & \textbf{Time}
& \textbf{Score} & \textbf{Time} \\
\midrule
EOH
& 58.05 & 1.15 & 314.20 & 3.54 & 58.25 & 0.69 & 96.46 & 0.56 \\
FunSearch
& 57.47 & 1.23 & 276.28 & 3.94 & 58.92 & 0.67 & 95.13 & 0.51 \\
EvoPrompting
& 58.01 & 0.28 & 488.65 & 0.47 & 61.62 & 1.21 & 96.90 & 0.13 \\
OpenEvolve
& 57.10 & 0.29 & 377.55 & 0.73 & 61.78 & 0.21 & 96.90 & 0.13 \\
\method
& \textbf{59.57} & \textbf{0.11}
& \textbf{500.00} & \textbf{0.04}
& \textbf{62.79} & 0.52
& \textbf{97.35} & 0.19 \\
\bottomrule
\end{tabular}
\caption{Raw task performance and end-to-end NAS time (GPU-hours) used to
construct the cross-task efficiency--performance plot.}
\label{tab:cross_task_raw}
\end{table*}

\paragraph{Aggregation.}

For method \(m\) on benchmark \(t\), task performance is normalized by
the best observed result:
\begin{equation}
    \widetilde{s}_{m,t}
    =
    \frac{s_{m,t}}{\max_{m'}s_{m',t}}.
\end{equation}
The vertical coordinate is the arithmetic mean:
\begin{equation}
    P_m
    =
    \frac{100}{|\mathcal{T}|}
    \sum_{t\in\mathcal{T}}\widetilde{s}_{m,t}.
\end{equation}

End-to-end NAS GPU hours is normalized by the fastest method on each task:
\begin{equation}
    \widetilde{\tau}_{m,t}
    =
    \frac{\tau_{m,t}}{\min_{m'}\tau_{m',t}}.
\end{equation}
Because these quantities are multiplicative ratios, the horizontal
coordinate is their geometric mean:
\begin{equation}
    C_m
    =
    \exp\left(
        \frac{1}{|\mathcal{T}|}
        \sum_{t\in\mathcal{T}}
        \log\widetilde{\tau}_{m,t}
    \right).
\end{equation}
Higher \(P_m\) and lower \(C_m\) are preferred.

\begin{center}
\scriptsize
\begin{minipage}{0.98\columnwidth}
\centering
\begin{tabular}{lrr}
\toprule
\textbf{Method}
& \textbf{Mean Relative Performance}
& \textbf{Relative NAS GPU hours} \\
\midrule
EOH & 88.036 & 10.697 \\
FunSearch & 85.822 & 10.836 \\
OpenEvolve & 92.323 & 2.634 \\
EvoPrompting & 98.196 & 3.623 \\
\method & \textbf{100.000} & \textbf{1.379} \\
\bottomrule
\end{tabular}
\captionof{table}{Aggregated coordinates used in the cross-task
efficiency--performance plot.}
\label{tab:cross_task_aggregate}
\end{minipage}
\end{center}

\section{Full CLRS Results}
\label{sec:clrs_results}

All methods except EvoPrompting are evolved on CLRS-DFS under the same
experimental protocol and evaluated zero-shot on all 30 tasks.
EvoPrompting uses its reported per-task results.

The aggregate results are reported in Table~\ref{tab:clrs_aggregate}; Tables~\ref{tab:clrs_accuracy}--\ref{tab:clrs_accuracy_part3} provide per-task accuracy, and Tables~\ref{tab:clrs_size}--\ref{tab:clrs_size_part3} provide per-task parameter counts.

\begin{table*}[t]
\centering
\small
\begin{tabular}{lrrrrr}
\toprule
\textbf{Method}
& \textbf{Model Size}
& \textbf{DFS Acc.}
& \textbf{Avg. Acc.}
& \textbf{Avg. w/o DFS}
& \textbf{Wins} \\
\midrule
EvoPrompting
& 478.2K & 68.14 & 80.88 & 81.32 & 2/30 \\
Triplet-GMPNN
& 467.3K & 47.79 & 75.98 & 76.95 & 1/30 \\
OpenEvolve
& 498.6K & 32.54 & 62.69 & 63.73 & 1/30 \\
Spark
& 470.2K & 83.74 & 83.91 & 83.92 & 5/30 \\
OpenEvolve+\method
& 467.4K & \textbf{94.68} & \textbf{89.21}
& \textbf{89.03} & \textbf{22/30} \\
\bottomrule
\end{tabular}
\caption{Aggregate CLRS results. Ties are counted as wins for every tied
method, so the win counts may sum to more than 30.}
\label{tab:clrs_aggregate}
\end{table*}

\begin{table*}[t]
\centering
\small
\setlength{\tabcolsep}{4pt}
\begin{tabular}{p{0.31\textwidth}ccccc}
\toprule
\textbf{Task} & \textbf{Initial} & \textbf{OpenEvolve} & \textbf{EvoPrompting} & \textbf{Spark} & \textbf{\method} \\
\midrule
articulation\_points & 88.32 & 61.32 & 93.46 & 97.90 & \textbf{99.40} \\
activity\_selector & 95.18 & 95.43 & 95.05 & 98.03 & \textbf{98.90} \\
bellman\_ford & 97.39 & 95.31 & 97.50 & 96.04 & \textbf{99.27} \\
bfs & 99.73 & \textbf{100.00} & 99.99 & 70.56 & \textbf{100.00} \\
binary\_search & 77.58 & 39.31 & 77.98 & 87.60 & \textbf{94.19} \\
bridges & 93.99 & 79.68 & 97.57 & 98.15 & \textbf{100.00} \\
bubble\_sort & 67.68 & 85.45 & 88.87 & \textbf{98.12} & 97.07 \\
dag\_shortest\_paths & 98.19 & 96.97 & 98.01 & 97.80 & \textbf{100.00} \\
dfs & 47.79 & 32.54 & 68.14 & 83.74 & \textbf{94.68} \\
dijkstra & 96.05 & 95.85 & 97.30 & 99.22 & \textbf{99.51} \\
\bottomrule
\end{tabular}
\caption{Per-task accuracy (\%) on the 30 CLRS tasks.}
\label{tab:clrs_accuracy}
\end{table*}

\begin{table*}[t]
\centering
\small
\setlength{\tabcolsep}{4pt}
\begin{tabular}{p{0.31\textwidth}ccccc}
\toprule
\textbf{Task} & \textbf{Initial} & \textbf{OpenEvolve} & \textbf{EvoPrompting} & \textbf{Spark} & \textbf{\method} \\
\midrule
find\_maximum\_subarray\_kadane & 76.36 & 77.05 & 75.35 & \textbf{85.64} & 82.52 \\
floyd\_warshall & 48.52 & 11.24 & \textbf{61.43} & 32.91 & 35.75 \\
graham\_scan & 93.62 & 83.29 & 93.76 & 96.36 & \textbf{97.60} \\
heapsort & 31.04 & 16.31 & 69.90 & 75.63 & \textbf{94.82} \\
insertion\_sort & 78.14 & 25.78 & 89.47 & 97.51 & \textbf{98.39} \\
jarvis\_march & 91.01 & 36.42 & 90.36 & \textbf{94.88} & 94.50 \\
kmp\_matcher & 19.51 & 10.09 & 16.29 & 14.94 & \textbf{52.00} \\
lcs\_length & 80.51 & 87.53 & 85.75 & 87.60 & \textbf{88.44} \\
matrix\_chain\_order & \textbf{91.68} & 80.84 & 90.77 & 87.29 & 89.88 \\
minimum & 97.78 & 98.10 & 98.40 & 99.07 & \textbf{99.51} \\
\bottomrule
\end{tabular}
\caption{Per-task accuracy (\%) on the 30 CLRS tasks (continued, Part 2 of 3).}
\label{tab:clrs_accuracy_part2}
\end{table*}

\begin{table*}[t]
\centering
\small
\setlength{\tabcolsep}{4pt}
\begin{tabular}{p{0.31\textwidth}ccccc}
\toprule
\textbf{Task} & \textbf{Initial} & \textbf{OpenEvolve} & \textbf{EvoPrompting} & \textbf{Spark} & \textbf{\method} \\
\midrule
mst\_kruskal & 89.80 & 76.57 & 91.47 & 86.23 & \textbf{91.96} \\
mst\_prim & 86.39 & 83.98 & 88.74 & 88.67 & \textbf{94.97} \\
naive\_string\_matcher & 78.67 & 8.93 & 79.77 & 77.27 & \textbf{100.00} \\
optimal\_bst & 73.77 & 71.73 & 78.66 & \textbf{96.14} & 84.84 \\
quickselect & 0.47 & 3.71 & 0.79 & 10.69 & \textbf{15.87} \\
quicksort & 64.64 & 16.26 & 85.23 & \textbf{97.80} & 96.58 \\
segments\_intersect & 97.64 & 96.85 & \textbf{98.15} & 97.07 & 98.09 \\
strongly\_connected\_components & 43.43 & 45.21 & 41.86 & 77.34 & \textbf{88.18} \\
task\_scheduling & 87.25 & 82.68 & 88.23 & 87.44 & \textbf{89.49} \\
topological\_sort & 87.27 & 86.31 & 88.12 & 99.72 & \textbf{100.00} \\
\midrule
\textbf{Average} & \textbf{75.98} & \textbf{62.69} & \textbf{80.88} & \textbf{83.91} & \textbf{89.21} \\
\bottomrule
\end{tabular}
\caption{Per-task accuracy (\%) on the 30 CLRS tasks (continued, Part 3 of 3).}
\label{tab:clrs_accuracy_part3}
\end{table*}

\begin{table*}[t]
\centering
\small
\setlength{\tabcolsep}{4pt}
\begin{tabular}{p{0.31\textwidth}ccccc}
\toprule
\textbf{Task} & \textbf{Initial} & \textbf{OpenEvolve} & \textbf{EvoPrompting} & \textbf{Spark} & \textbf{\method} \\
\midrule
articulation\_points & 531913 & 563259 & 497969 & 534880 & 532008 \\
activity\_selector & 262204 & 293550 & 262204 & 265171 & 262299 \\
bellman\_ford & 524604 & 555950 & 568660 & 527571 & 524699 \\
bfs & 523963 & 555309 & 522931 & 526930 & 524058 \\
binary\_search & 262204 & 293550 & 262204 & 265171 & 262299 \\
bridges & 532556 & 563902 & 576612 & 535523 & 532651 \\
bubble\_sort & 524477 & 555823 & 568533 & 527444 & 524572 \\
dag\_shortest\_paths & 793287 & 824633 & 793287 & 796254 & 793382 \\
dfs & 661190 & 692536 & 660158 & 664157 & 661285 \\
dijkstra & 525886 & 557232 & 524854 & 528853 & 525981 \\
\bottomrule
\end{tabular}
\caption{Per-task model size, measured by total parameter count, on the 30 CLRS tasks.}
\label{tab:clrs_size}
\end{table*}

\begin{table*}[t]
\centering
\small
\setlength{\tabcolsep}{4pt}
\begin{tabular}{p{0.31\textwidth}ccccc}
\toprule
\textbf{Task} & \textbf{Initial} & \textbf{OpenEvolve} & \textbf{EvoPrompting} & \textbf{Spark} & \textbf{\method} \\
\midrule
find\_maximum\_subarray\_kadane & 264514 & 295860 & 261290 & 267481 & 264609 \\
floyd\_warshall & 625089 & 656435 & 669145 & 628056 & 625184 \\
graham\_scan & 398409 & 429755 & 397377 & 401376 & 398504 \\
heapsort & 659654 & 691000 & 703710 & 662621 & 659749 \\
insertion\_sort & 524477 & 555823 & 523445 & 527444 & 524572 \\
jarvis\_march & 264898 & 296244 & 308954 & 267865 & 264993 \\
kmp\_matcher & 398021 & 729367 & 396989 & 400988 & 398116 \\
lcs\_length & 270419 & 301765 & 270419 & 273386 & 270514 \\
matrix\_chain\_order & 624448 & 655794 & 624448 & 627415 & 624543 \\
minimum & 261307 & 292653 & 260275 & 264274 & 261402 \\
\bottomrule
\end{tabular}
\caption{Per-task model size, measured by total parameter count, on the 30 CLRS tasks (continued, Part 2 of 3).}
\label{tab:clrs_size_part2}
\end{table*}

\begin{table*}[t]
\centering
\small
\setlength{\tabcolsep}{4pt}
\begin{tabular}{p{0.31\textwidth}ccccc}
\toprule
\textbf{Task} & \textbf{Initial} & \textbf{OpenEvolve} & \textbf{EvoPrompting} & \textbf{Spark} & \textbf{\method} \\
\midrule
mst\_kruskal & 399691 & 431017 & 443747 & 402658 & 399786 \\
mst\_prim & 525886 & 557232 & 569942 & 528853 & 525981 \\
naive\_string\_matcher & 262588 & 293934 & 259364 & 265555 & 262683 \\
optimal\_bst & 625987 & 357333 & 624955 & 628954 & 626082 \\
quickselect & 395714 & 427060 & 377130 & 398681 & 395809 \\
quicksort & 525759 & 557105 & 524727 & 528726 & 525854 \\
segments\_intersect & 263359 & 294705 & 262327 & 266326 & 263454 \\
strongly\_connected\_components & 663243 & 694589 & 707299 & 666210 & 663338 \\
task\_scheduling & 262333 & 293679 & 262333 & 265300 & 262428 \\
topological\_sort & 660164 & 691510 & 660164 & 663131 & 660259 \\
\midrule
\textbf{Average} & \textbf{467274.80} & \textbf{498620.13} & \textbf{478181.73} & \textbf{470241.80} & \textbf{467369.80} \\
\bottomrule
\end{tabular}
\caption{Per-task model size, measured by total parameter count, on the 30 CLRS tasks (continued, Part 3 of 3).}
\label{tab:clrs_size_part3}
\end{table*}




\end{document}